\documentclass[runningheads]{llncs}

\usepackage{eccv}
\usepackage{eccvabbrv}
\usepackage{graphicx}
\usepackage{booktabs}
\usepackage{amsmath,amssymb}
\usepackage{xcolor}
\usepackage[accsupp]{axessibility}
\usepackage{hyperref}

\def\ie{\textit{i.e.}}

\title{PhysChoreo: Physics-Controllable Video Generation with Part-Aware Semantic Grounding}

\titlerunning{PhysChoreo}

\author{
Haoze Zhang\inst{1} \and
Tianyu Huang\inst{1} \and
Zichen Wan\inst{1} \and
Xiaowei Jin\inst{1} \and
Hongzhi Zhang\inst{1} \and
Hui Li\inst{1} \and
Wangmeng Zuo\inst{1}\thanks{Corresponding author. Email: wmzuo@hit.edu.cn}
}

\authorrunning{H. Zhang et al.}

\institute{
\textsuperscript{1}Harbin Institute of Technology, China
\\
\email{kakaka0521@outlook.com}
\\
Project: \url{https://kakaka0521.github.io/physchoreo_web}
}
\begin{document}

\maketitle

\begin{abstract}
While recent video generation models have achieved significant visual fidelity, they often suffer from the lack of explicit physical controllability and plausibility. To address this, some recent studies attempted to guide video generation with physics-based rendering. However, these methods face inherent challenges in accurately modeling complex physical properties and effectively controlling the resulting physical behavior over extended temporal sequences. In this work, we introduce PhysChoreo, a framework that can generate videos with diverse controllability and physical realism from a single image. Our method consists of two stages: first, it estimates the static initial physical properties of all objects in the image through part-aware physical property reconstruction. Then, through temporally instructed and physically editable simulation, it synthesizes high-quality videos with rich dynamic behaviors and physical realism. Experimental results show that PhysChoreo can generate videos with rich behaviors and physical realism, outperforming state-of-the-art methods on multiple evaluation metrics.

\end{abstract}

\begin{figure*}[t]
    \centering
    \includegraphics[width=1\textwidth]{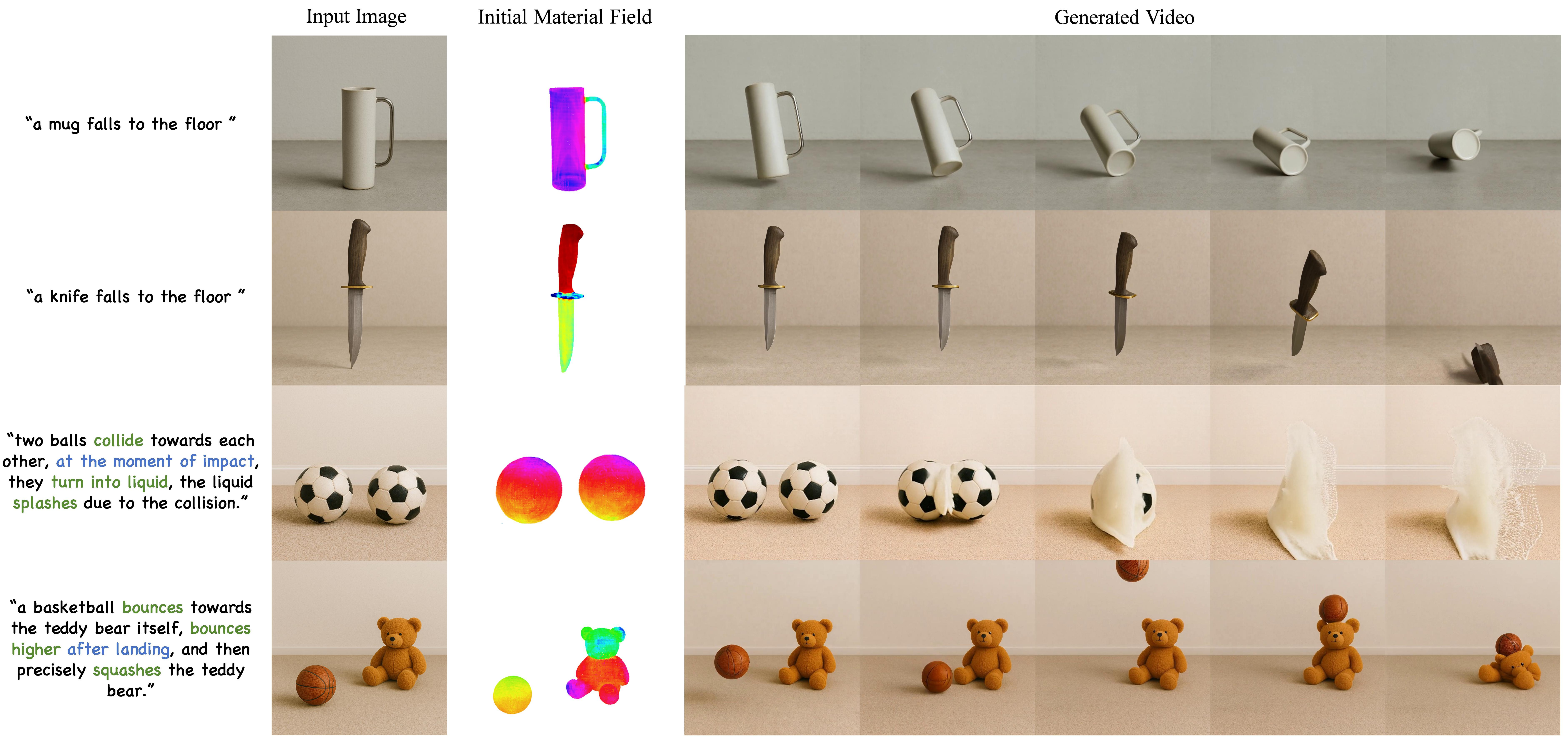}
    \caption{We propose PhysChoreo, a framework for controllable image-to-video generation. PhysChoreo reconstructs the material field of objects from a single image and generates physically realistic and dynamically rich videos. In (a) and (b), physically realistic dynamics are generated based on reconstructed physical properties. In (c) and (d), by controlling physical properties during generation, more cinematic videos can be produced while maintaining physical realism.}
    \label{fig:intro}
\end{figure*}

\section{Introduction}
Recent advances in video generation~\cite{blattmann2023stable,yang2024cogvideox, wang2025lavie, he2022latent,chen2023videocrafter1opendiffusionmodels} have significantly improved visual fidelity and spatiotemporal consistency, yet the physical realism of generated content remains limited~\cite{kang2024far,wiedemer2025video}. Contemporary models primarily scale up by learning from vast amounts of data to capture physical phenomena, rather than understanding the underlying principles that govern how objects respond to forces and material constraints. As a consequence, generated videos often fail to accurately present real-world behavior, particularly in complex or even counterfactual scenarios. These limitations reveal that current models highly rely on simplistic imitation of motion patterns, lacking causal reasoning about physical rules.

To produce physically plausible outcomes, recent works~\cite{xie2024physgaussian,lin2025omniphysgs, zhang2024physdreamer, huang2025dreamphysics,tan2024physmotion} have attempted to incorporate physical simulation into the generation procedure. Specifically, these methods employ finite element simulation to compute the motion state of the target objects, which is then used as the condition to render or generate corresponding dynamic effects. Since the process of physical simulation is deterministic, the key challenges become: (1) modeling the physics fields of the target scene, and (2) controlling the generated content with simulated results. However, existing methods have not effectively addressed these two issues. On the one hand, physics optimization approaches~\cite{lin2025omniphysgs, zhang2024physdreamer, huang2025dreamphysics} typically impose coarse prediction on single objects, which fails to meet the modeling requirements of real-world simulation. On the other hand, simulators~\cite{xie2024physgaussian,jiang2016material,muller2020detailed} used in these methods are not as flexible in control as video models, resulting in limited simulation modes.

We address these limitations with \textbf{PhysChoreo}, a physics-based video generation framework that consists of \emph{part-aware semantic physics} prediction and \emph{physics-editable control}. The key idea for \emph{part-aware semantic physics} is to align textual semantics with detailed 3D part structures. For each part in an object, both the material model and calibrated continuous physical quantities are estimated with our pre-trained predictor. Consequently, a part-level physics supervision strategy is proposed, which tightly couples semantics, geometry, and physics at fine granularity, enabling our predictor to reconcile physical interpretability and controllability. Beyond static physics estimation, we integrate physics-based simulation~\cite{hu2019difftaichi,hu2019taichi,hu2021quantaichi} with \emph{physics-editable control} directly injected into the generative process. Our control allows the model to modulate external forces and material attributes over time, so that objects respond in a physically plausible manner throughout a sequence. This formulation converts high-level instructions into temporally consistent physical changes, making counterfactual interventions such as liquefying upon collision, collapsing upon landing, counter-intuitive bounces, both natural and verifiable.

To support part-level physics modeling, we construct a fine-grained \emph{text--part--physics} dataset that covers common materials and canonical part relations across a variety of objects. For each instance, the dataset provides a global textual description and part-level phrase prompts that capture functional or material cues. Both point-level and part-level labels are annotated, including discrete material categories together with key continuous physical quantities, \ie, Young’s modulus $E$, Poisson’s ratio $\nu$, and density $\rho$.
This resource serves as a unified benchmark for training and evaluating per-part physics estimation.

We conduct extensive experiments to evaluate the prediction of physical properties and the quality of physics-controllable video generation. For \emph{part-level physics} prediction, the proposed method advances the state-of-the-art in continuous parameter regression, maintaining robust generalization across different categories and scenarios. For \emph{physics-controllable} video generation, we used more complex and diverse instructions to generate videos with various visual effects, and systematically evaluate the physical realism, instruction following, and visual quality of the generated videos, showing that PhysChoreo can faithfully realize instruction-driven changes while maintaining high visual fidelity and temporal coherence.

Our contributions can be summarized as:
\begin{itemize}
    \item We release a text–part–physics dataset with consistent semantic, geometric, and physical annotations, establishing a reusable benchmark for training and evaluation.
    \item We propose to predict \emph{part-level} semantic physics modeling via soft assignment and hierarchical cross-attention, which is supervised by multiple physics constraints.
    \item We propose \emph{physics-editable control} in dynamics simulation, which formalizes editing as temporally continuous intervention, enabling flexible, fine-grained modulation and interaction.
    \item We introduce PhysChoreo, a physics-based video generation framework that reconstructs physical properties and generates diverse controllable dynamic sequences.
\end{itemize}
\section{Related Work}
\noindent\textbf{Physical Property Estimation.} Existing methods can be broadly divided into two categories: runtime optimization and direct estimation. Runtime optimization methods~\cite{huang2025dreamphysics,lin2025omniphysgs,zhang2024physdreamer} gradually optimize physical properties through iterative processes guided by rendered videos and diffusion models. However, these methods can't supervise fine-grained physical modeling and incur significant time overhead, making them difficult to use for direct video generation. Among direct estimation methods, NeRF2Physics~\cite{zhai2024physical} and PUGS~\cite{shuai2025pugszeroshotphysicalunderstanding} use multi-view images to enable VLMs to infer physical properties. However, they both rely on generated 2D features and cannot be applied to general 3D representations. Pixie~\cite{le2025pixie} injects features in 3D through multi-view images and performs direct estimation, but it also cannot be directly applied to general 3D representations and can only learn static features rather than their distributions.

\noindent\textbf{Physics-Based Video Generation.} Existing methods utilize physical simulators to generate physics-based videos. Some methods, such as PhysGaussian~\cite{xie2024physgaussian} and subsequent work~\cite{zhang2024physdreamer,huang2025dreamphysics,lin2025omniphysgs,liu2025physflowunleashingpotentialmultimodal,mittal2025uniphylearningunifiedconstitutive}, reconstruct scene representations from multi-view images, simulate these representations, and then render videos. However, these methods rely on high-quality 3D reconstruction. Recent methods~\cite{liu2024physgen,chen2025physgen3d,tan2024physmotion} consider generating dynamics using physical simulators from a single image and synthesizing them, while others~\cite{xie2025physanimatorphysicsguidedgenerativecartoon,li2025wonderplay,gillman2025forcepromptingvideogeneration} consider using dynamics as guidance to assist video models in generation. PhysCtrl~\cite{physctrl2025} directly uses diffusion models to generate dynamics and then employs video models for generation. However, these methods depend on manual physics parameter settings and lack fine-grained temporal control, resulting in short and monotonous generated dynamics.

\noindent\textbf{Controllable Generative Video Model.} Generative video models are trained on large-scale data to acquire the ability to generate videos and can produce high-quality videos~\cite{deepmind2025veo,yang2024cogvideox,xing2024dynamicrafter,chen2024videocrafter2overcomingdatalimitations,chen2023videocrafter1opendiffusionmodels}, but they often lack controllability. Recent methods have explored various approaches for controllable video generation, where optical flow~\cite{ni2023conditionalimagetovideogenerationlatent,liang2024movideomotionawarevideogeneration}, motion~\cite{gu2025diffusion,li2025magicmotioncontrollablevideogeneration}, and others can serve as control conditions to guide generation. However, they still lack physical plausibility and controllability. A key aspect of our work is generating rich dynamics to guide the generation process, aiming to achieve physically realistic videos.
\section{Method}
In this section, we introduce \textbf{PhysChoreo}, a novel framework for physics-controllable video generation.  Given 3D objects reconstructed from a single image, PhysChoreo conducts a part-aware physics prediction via soft assignment and hierarchical cross-attention (See Sec.~\ref{sec:31}). A part-level physics supervision is then proposed to facilitate the training of the physics predictor in terms of local smoothness and prompt guidance (See Sec.~\ref{sec:32}).
Finally, the predicted physics field drives controllable simulators and video model to produce editable, physically consistent videos (See Sec.~\ref{sec:33}).
\begin{figure*}[t]
    \centering
    \includegraphics[width=1\textwidth]{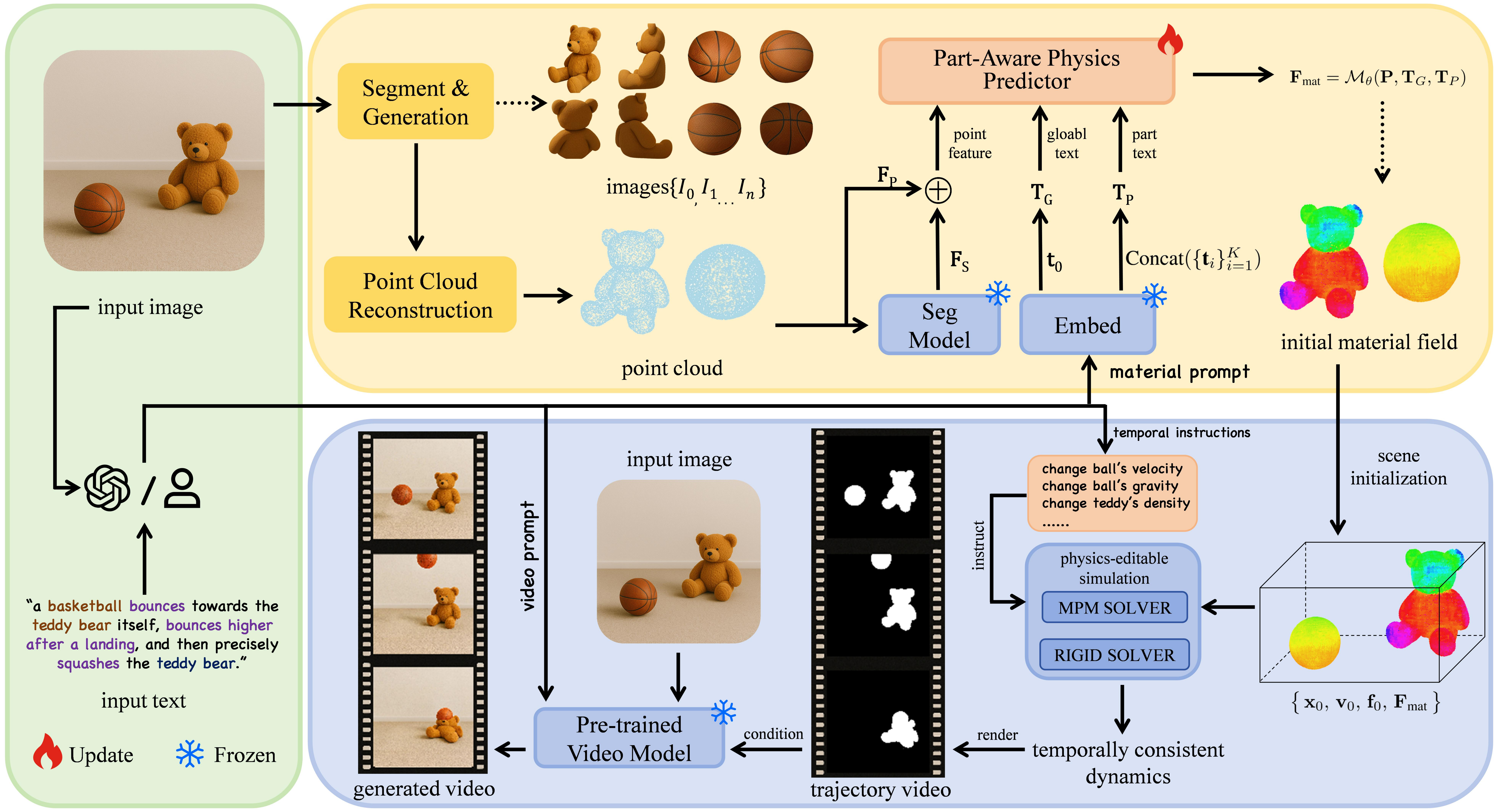}
\caption{Overview of our pipeline. Given the input image and text prompt, we first reconstruct the initial material field of each object from the image. Then we generate the scene's trajectory video based on a physics-editable simulator with temporal instructions, and finally the trajectory video is used as conditional control to guide the generation of generative video model.}
    \label{fig:pipeline}
\end{figure*}
\begin{figure*}[t]
    \centering
    \includegraphics[width=1\textwidth]{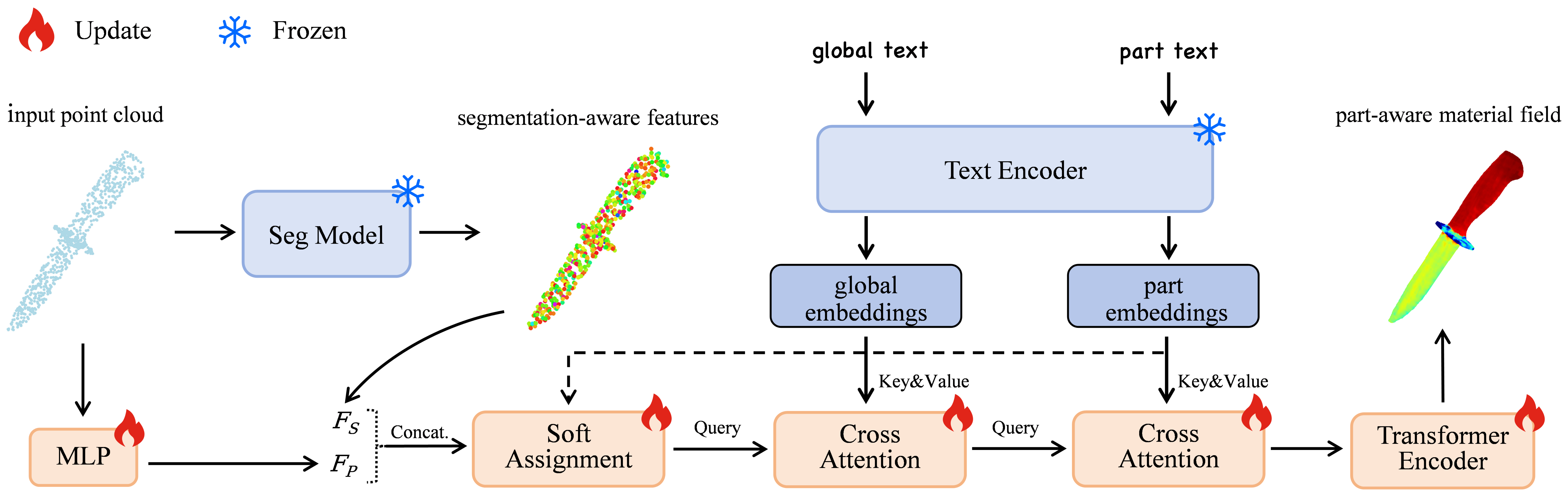}
\caption{Overview of our model design. We first use a fused feature from point positional feature and segmentation prior. Afterward, we use a soft assignment to preliminarily display the injected part-level features, then perform fine-grained text-level adjustments through a hierarchical cross-attention stage, and finally obtain part-aware material field features via a transformer encoder.}
    \label{fig:model}
\end{figure*}
\subsection{Part-Aware Physics Reconstruction}
\label{sec:31}
Given an input image $I$, our intention is to reconstruct the corresponding scene and assign reasonable physical properties to all objects in the scene. To this end, we segment all the instances in $I$ and then reconstruct a dense triangular mesh for each instance with InstantMesh~\cite{xu2024instantmesh}. We obtain the point cloud representation $P_i\in \mathbb{R}^{N\times3}$ by uniformly sampling $N$ points on the surface of the $i$-th mesh, which is the target of the following physics prediction.

To specify the material, roughness, and other related physical properties of $P_i$, an input text prompt $y_0$ is provided, which is then encoded into text embeddings $\textbf{t}_{0}\in\mathbb{R}^{d_t}$ with CLIP~\cite{radford2021learning}. Considering that one object may contain multiple materials, part-level prompts $\{y_i\}_{i=1}^K$($K$ denotes the number of part prompts) are optionally provided to describe the detailed properties for different parts of $P_i$, which are also embedded as $\{\textbf{t}_{i}\}_{i=1}^K$. Accordingly, we extract global point features $\textbf{F}_P\in \mathbb{R}^{N\times d_P}$ and part-level semantic features $\textbf{F}_S\in \mathbb{R}^{N\times d_S}$ with a lightweight MLP and a pre-trained part segmentation encoder~\cite{partfield2025,liu2023partslip,yang2024sampart3d,Su_2026_CVPR}, respectively. In order to convert text instructions into a point-based physics field that is globally coherent and locally editable, we propose soft assignment to align the feature dimension of point cloud with part-level text embeddings and hierarchical cross-attention to combine multi-granularity text embeddings with point features.

\noindent\textbf{Soft Assignment.}
Our aim is to inject part-level semantics into point features in a way that is editable, stable, and interpretable, i.e., each point should carry an explicit distribution over part prompts while its geometric signal is preserved. Therefore, we align points and prompts in a shared latent space and use the resulting weights to mix prompt “values” back into the point stream via a residual update. Concretely, we formulate an additive refinement of point features $\textbf{H}=\text{Concat}(\textbf{F}_S,\textbf{F}_P)\!\in\!\mathbb{R}^{N\times d}$ with part-prompt embeddings $\textbf{T}=\text{Concat}(\{\textbf{t}_i\}_{i=1}^K)\!\in\!\mathbb{R}^{K\times d_t}$ as:
\begin{equation}
A = \operatorname{softmax}_{\text{row}}((\textbf{H} \Phi)\,(\textbf{T} \Psi)^{\top}) \in \mathbb{R}^{N\times K},
\end{equation}
\begin{equation}
\hat{\textbf{H}}= \textbf{H} + A\,(\textbf{T} W) \in \mathbb{R}^{N\times d},
\end{equation}
where $\Phi\!\in\!\mathbb{R}^{d\times d_a}$ and $\Psi\!\in\!\mathbb{R}^{d_t\times d_a}$ project points and prompts to the alignment space with dimension $d_a$, $W\!\in\!\mathbb{R}^{d_t\times d}$ maps prompt “values” to the point-feature space, and $\operatorname{softmax}_{\text{row}}$ normalizes across the $K$ prompts for each point so that each row of $A$ sums to $1$. Algebraically, this block is equivalent to a \emph{single-head cross-attention}, but we deliberately adopt this minimal form so that the assignment distribution $A$ remains directly interpretable and its logits $S$ can be explicitly supervised by $L_{\text{assign}}$ in ~\ref{sec:32}.

\noindent\textbf{Hierarchical Cross-Attention.}
To guide the point stream with global semantics before part-level details refine it, we propose a hierarchical cross-attention mechanism to support the interaction of the point stream and the multi-level text stream. Let $\textbf{T}_0=[\textbf{t}_0]\in\mathbb{R}^{1\times d_t}$ be the global text token and $\textbf{T}$ be the part-level token. Queries always come from the point stream, while keys/values come from text:
\begin{equation}
\textbf{H}_g = \mathrm{MHA}(\hat{\textbf{H}},\, \textbf{T}_0) + \hat{\textbf{H}},
\end{equation}
\begin{equation}
\textbf{H}_p = \mathrm{MHA}(\textbf{H}_g,\, \textbf{T}) + \textbf{H}_g,
\end{equation}
\noindent The first stage uses a single global token to impose scene-level consistency on all points, acting as a coarse “style” conditioner that stabilizes subsequent language injection. The second stage then attends to the part tokens, sharpening locality and disentangling part-specific effects without overriding the coarse global guidance. Placing the global stage before the part stage reduces competition between tokens and empirically improves convergence and editability.

To aggregate non-local context while remaining permutation-invariant over points, the part-conditioned features $\textbf{H}_p$ are encoded by a set Transformer encoder~\cite{zhao2021point} to produce point embeddings $Z=Transformer(\textbf{H}_p)\in\mathbb{R}^{N\times d_z}$. Then, standard point-wise heads decode (i) class logits followed by a softmax to yield per-point material model class probabilities $\hat{Y}\in\mathbb{R}^{N\times C}$ and (ii) continuous material parameters $\hat{M}\in\mathbb{R}^{N\times q}$. For each point $i$, the material field is represented as
\begin{equation}
\label{eq:material_field_pointwise}
M_{\mathrm{pred}}(x_i)=\Big(\operatorname*{arg\,max}_{c\in\{1,\dots,C\}}\hat{Y}_{i,c},\ \hat{M}_{i,:}\Big).
\end{equation}

\subsection{Part-Level Physics Supervision}
\label{sec:32}
We couple point-wise supervision with structural priors so that the learned material field is
(i) correct at each point, (ii) smooth \emph{within} semantic parts, and (iii) aligned with the part prompts used for conditioning.
Let $\{x_i\}_{i=1}^{N}$ be sampled 3D points on the object surface/volume, where $x_i\in\mathbb{R}^3$ and $N$ is the number of supervised points.
For each point $i$, the network predicts continuous material parameters $\hat{m}_i$ (e.g., $(\hat{E}_i,\hat{\nu}_i,\hat{\rho}_i)$) and
a material-model class distribution $\hat{y}^{\mathrm{cls}}_i\in\Delta^{C-1}$ over $C$ classes.
Ground truth is given by $m_i$ and a class label $y^{\mathrm{cls}}_i\in\{1,\ldots,C\}$.
We write $\operatorname{CE}(\hat{y}^{\mathrm{cls}}_i,y^{\mathrm{cls}}_i)=-\log \hat{y}^{\mathrm{cls}}_{i,y^{\mathrm{cls}}_i}$ and use a smooth regression loss
$\ell(\hat{m}_i,m_i)$ on normalized targets.

\noindent\textbf{Task Supervision.}
This term anchors the semantic identity and calibrates physical magnitudes on a per-point basis:
\begin{equation}
\label{eq:task_loss_final}
    L_{\text{task}}=\frac{1}{N}\sum_{i=1}^{N}\Big[
        \lambda_{\text{reg}}\cdot\ell\big(\hat{m}_i,\,m_i\big)
        +\lambda_{\text{cls}}\cdot\operatorname{CE}\big(\hat{y}^{\mathrm{cls}}_i,\,y^{\mathrm{cls}}_i\big)
    \Big],
\end{equation}

\noindent\textbf{Wave Continuity.}
In both the physical world and simulation environments, the propagation speed of elastic waves, as determined by material parameters such as
Young's modulus, Poisson's ratio, and density, is an intrinsic reflection of the material's mechanical characteristics~\cite{PNG2025103431}.
To ensure the learning of relationships between physical properties and spatial continuity, we regularize the predicted longitudinal and shear wave speeds.
Given material parameters $(E_i,\nu_i,\rho_i)$ at point $x_i$, we compute
\begin{equation}
\label{eq:wavespeed_def}
c_p(x_i) =
\sqrt{
    \frac{E_i (1 - \nu_i)}
    {\rho_i (1 + \nu_i)(1 - 2\nu_i)}
},
\qquad
c_s(x_i) =
\sqrt{
    \frac{E_i}
    {2\rho_i (1 + \nu_i)}
},
\end{equation}
where $E_i$ is Young's modulus, $\nu_i$ is Poisson's ratio, and $\rho_i$ is density.

We then approximate the continuous smoothness penalty $\int \|\nabla c_p(x)\|_2^2+\|\nabla c_s(x)\|_2^2\,dx$ on sampled points using a within-part neighborhood graph.
Let $\mathcal{N}(i)$ be the $k$-nearest-neighbor set of point $i$ in 3D, and let $\mathrm{part}(i)$ be its ground-truth semantic part label.
We define the within-part edge set
$\mathcal{E}=\{(i,j)\mid j\in\mathcal{N}(i),\, \mathrm{part}(i)=\mathrm{part}(j)\}$ and minimize
\begin{equation}
\label{eq:smooth_discrete}
L_{\text{smooth}} =
\frac{1}{|\mathcal{E}|}\sum_{(i,j)\in\mathcal{E}}
w_{ij}\Big(
\big(c_p(x_i)-c_p(x_j)\big)^2
+
\big(c_s(x_i)-c_s(x_j)\big)^2
\Big),
\end{equation}
where $|\mathcal{E}|$ is the number of within-part edges and $w_{ij}=(\|x_i-x_j\|_2^2)^{-1}$ is a standard discretization weight.
This loss penalizes spatial discontinuities \emph{within} semantic parts and prevents non-physical jumps in the object.

\noindent\textbf{Contrastive Regularization.}
Beyond the within-part continuity enforced by $L_{\text{smooth}}$, we add a physically motivated contrast to keep parts separable near interfaces.
From $(E_i,\nu_i)$ we compute the shear modulus $\mu_i=\frac{E_i}{2(1+\nu_i)}$ and the bulk modulus $K_i=\frac{E_i}{3(1-2\nu_i)}$.
We form a log-domain, $\ell_2$-normalized embedding
\begin{equation}
\label{eq:phys_embed}
e_i=\mathrm{norm}\big([\log \mu_i,\ \log K_i]\big),
\end{equation}
where $\mathrm{norm}(\cdot)$ denotes $\ell_2$ normalization.
For an anchor $i$, we choose a positive $p\in P(i)$ and a negative $n\in N(i)$, where
$P(i)=\{j \mid \mathrm{part}(j)=\mathrm{part}(i)\}$ and $N(i)=\{j \mid \mathrm{part}(j)\neq \mathrm{part}(i)\}$, and minimize the triplet loss
\begin{equation}
\label{eq:triplet_phys}
L_{\text{con}}
=\frac{1}{|\mathcal{T}|}\sum_{(i,p,n)\in\mathcal{T}}
\max\!\big(0,\ \|e_i{-}e_p\|_2^2 - \|e_i{-}e_n\|_2^2 + m\big),
\end{equation}
where $\mathcal{T}$ is the set of sampled triplets and $m>0$ is the margin.
Intuitively, $L_{\text{smooth}}$ enforces within-part propagation consistency, while $L_{\text{con}}$ preserves inter-part separability in elastic response.

\noindent\textbf{Prompt--Part Assignment.}
Soft assignment provides an interpretable point-to-prompt distribution.
Let $s_{ik}$ denote the similarity logit between point $i$ and prompt $k\in\{1,\ldots,K\}$, and define
$a_i=\operatorname{softmax}_k(s_{ik})\in\Delta^{K-1}$ as the assignment distribution.
We encourage this distribution to match the ground-truth part label via cross-entropy:
\begin{equation}
\label{eq:assign_loss}
L_{\text{assign}}
=\frac{1}{N}\sum_{i=1}^{N}
\operatorname{CE}\!\big(a_i,\ \pi(\mathrm{part}(i))\big),
\end{equation}
where $\pi(\cdot)$ maps a part label to its prompt index and $\operatorname{CE}(a_i,\pi)=-\log a_{i,\pi}$.
This couples the language interface to geometry, improving interpretability and editability without altering the task-loss pathway.

\noindent\textbf{Overall Objective.}
The overall loss is a weighted sum of the above terms:
\begin{equation}
\label{eq:overall_loss_final}
    L= L_{\text{task}}
      + \lambda_{\text{smooth}}\,L_{\text{smooth}}
      + \lambda_{\text{con}}\,L_{\text{con}}
      + \lambda_{\text{assign}}\,L_{\text{assign}},
\end{equation}
where $(\lambda_{\text{smooth}},\lambda_{\text{con}},\lambda_{\text{assign}})$ are tuned on a validation set.

\subsection{Physics-Editable Video Generation}
\label{sec:33}
Previous methods~\cite{liu2024physgen,tan2024physmotion} rely on manual initialization and only generate dynamics under initial conditions; as a result, the generated dynamics are short in duration and lack subsequent process control, which limits the diversity. Physical simulation of PhysChoreo enhances the controllability and diversity of dynamics by introducing a physics-editable dynamic simulation. Specifically, the simulator executes predefined, time-indexed action sequences during runtime, enabling complex effects without restarting the simulation. Afterward, the dynamic trajectories are fed to a pre-trained video model to generate.

\noindent\textbf{Physics-Editable Dynamics.}
We introduce an editable guide for MPM~\cite{jiang2016material} and Rigid~\cite{muller2020detailed} simulations that supports temporal control. Specifically, we maintain the physical properties of each object in the scene. The corresponding properties, including constitutive parameters like Young's modulus and density, external force fields like gravity and wind, and object momentum like velocity, can be controlled individually. To maintain physical plausibility under non-steady states, we apply continuity constraints to transitions and limit extreme values. For object/part $j$ at simulation step $n$, we smooth edited parameters
$\boldsymbol{\theta}_j^n=(E_j^n,\nu_j^n,\rho_j^n)$ toward the user target
$\tilde{\boldsymbol{\theta}}_j^n$ by
\begin{equation}
\boldsymbol{\theta}_{j}^{n+1}
=(1-\gamma_n)\boldsymbol{\theta}_{j}^{n}
+\gamma_n\tilde{\boldsymbol{\theta}}_{j}^{n},
\quad
\gamma_n=1-e^{-\Delta t/\tau_j}.
\end{equation}
where $\Delta t$ is the solver step, $\tau_j$ is the transition time constant, and $\gamma_n$ is the smoothing weight. We further bound stiffness by the MPM wave speed
$c_j^2=(\lambda_j+2\mu_j)/\rho_j$,
where
$\lambda_j=\frac{E_j\nu_j}{(1+\nu_j)(1-2\nu_j)}$ and
$\mu_j=\frac{E_j}{2(1+\nu_j)}$, using
\begin{equation}
s=\min\!\left(1,\frac{c_{\max}^2}{\max_j c_j^2+\epsilon}\right),
\quad
E'_j=sE_j .
\end{equation}
Thus edits are temporally continuous, and the shared scale $s$ preserves relative stiffness. This makes simulation both flexible and stable, achieving fine-grained control without resetting the scene. Through this temporal control, we can create diverse yet physically realistic visual effects. For example, by eliminating the density of internal particles, we can achieve hollow or deflating; by controlling the force field on different objects, we can achieve counter-intuitive motion or bullet time; and by changing the material model, we can achieve transformation.
All dynamic behaviors can be applied temporally, providing precision, controllability, and diversity for dynamics.

\noindent\textbf{Video Generation.}
Given the input image and text prompt, we first use Dust3r~\cite{wang2024dust3r} to estimate the positions and scales of all the objects in the scene, initializing the spatial information via point cloud reconstruction and part-level physics prediction. Considering that the reconstructed points primarily lie on the surface of objects, volumetric completion is required to obtain a solid representation for physical simulation. To this end, we employ a surface-to-interior propagation algorithm to generate particles for filling the interior of objects, which ensures seamless transitions from surface to interior. To assign the physical properties for filled points, we identify each interior particle's nearest surface particle using a k-nearest neighbors search and directly inherit the surface particle's properties to it. This process ensures seamless material transitions from object boundaries to interior volume during simulation. Finally, we use physics-editable simulators to produce point cloud motion trajectories, which are fed to a pre-trained video model~\cite{wan2025wan} as conditions to generate videos with physical realism.
\section{Dataset}
We introduce the dataset used to train and evaluate our prediction model. It is one of the largest point cloud datasets that couples part segmentation with global descriptions, and part-level physical properties with textual annotations.
\subsection{Dataset Pipeline}
Our pipeline can be divided into three steps. First, we clean the labels of the segmented 3D data, including merging labels of the same object that are over-segmented and deleting labels that will not be used for part descriptions. We input the rendering of the object and the object's category name into GPT-5~\cite{gpt5} for reasoning, guiding it to generate an overall object description with vague physical descriptions within physically reasonable ranges based on geometric features. Then, we feed the rendering and name of each part along with the overall object description into the VLM for further reasoning, generating textual descriptions and physical properties for each part. Finally, we refine annotations by manually defining constraints between materials and physical properties, where GLM-4.5~\cite{glm2024chatglm} is employed to calibrate the correspondence between materials and physical properties. We combine the reasoning output of both VLM and LLM, ensuring the correctness of annotations, while manually verifying 9\% of samples to guarantee quality.

\subsection{Dataset Overview}
We collect \textbf{9,580} samples that span 24 semantic categories with segmentation information from PartNet~\cite{mo2019partnet}. Each sample has an overall description, part-level physical properties for each part, including the real material, Young's modulus, density, and Poisson's ratio. Besides, to make the data more adaptable to simulation, we provide the mapped simulator material tag that corresponds to the initial real material description. Moreover, to increase the difficulty of training and evaluation, we deliberately introduce “counterfactual” labels for $5\%$ of the examples (e.g., a gelatinous blade, a metallic flower), thereby promoting the model’s ability to learn the associations between text and physics.
\section{Experiments}
In this section, we conduct extensive experimental comparisons on both the prediction of physical properties and the generation of physics-controllable videos. Ablation studies are further conducted to corroborate the effectiveness of our newly proposed modules.

\subsection{Implementation Details.}
We use GPT-5~\cite{gpt5} reasoning to split text into a global description, part prompts, solver instructions, as well as an input text prompt. For instance reconstruction, we use Grounded-Sam~\cite{ren2401grounded,liu2024grounding,kirillov2023segment} to segment instances from the image and use InstantMesh~\cite{xu2024instantmesh} to generate corresponding meshes. A pre-trained part segmentation encoder PartField~\cite{partfield2025} is then adopted to obtain 96-dim part-level semantics, concatenated with 96-dim embedding from a MLP, is conditioned with frozen CLIP~\cite{radford2021learning} 256-dim text features from both global description and part prompts; soft assignment is computed with $\tau=0.07$; the concatenated features are mapped to 512 dimensions, then pass through 8-head cross-attention and a 6-layer, 8-head transformer encoder. We predict continuous physical quantities and 6 material models for simulation. The model is trained on 2 A6000 GPUs, with a learning rate of $3\times10^{-4}$, batch size of 32, AdamW~\cite{loshchilov2017decoupled}, cosine decay~\cite{loshchilov2016sgdr}, respective weights $\lambda_{\text{reg}}=1,\lambda_{\text{cls}}=0.3,  \lambda_{\text{assign}}=0.1,\lambda_{\text{smooth}}=0.02,\lambda_{\text{con}}=5\times10^{-4}$, and patience-based early stopping on validation with $P=10$, $\Delta=10^{-4}$, checkpoint retained with training halts at epoch~41.
For video generation, we use Taichi-based~\cite{hu2019taichi} simulation and a pre-trained image-to-video model Wan2.2-Fun-5B-Control~\cite{wan2025wan}. PhysChoreo can be deployed on a single RTX 5090 GPU. The physics reconstruction and pose estimation stages take around 120 seconds, and the simulation takes about 30 seconds.
We choose GPT-5~\cite{gpt5} for multiple rounds of instruction optimization with complex dynamics to achieve better results.
Please refer to Suppl. for details of prompts, dataset, training particulars, and solvers.

\subsection{Physical Property Prediction}
\noindent\textbf{Baselines.}
We compare the predictor of PhysChoreo against a recently proposed multi-view prediction method Pixie~\cite{le2025pixie} and two large vision-language model methods NeRF2Physics~\cite{zhai2024physical} and PUGS~\cite{shuai2025pugszeroshotphysicalunderstanding}. Pixie renders the multiple views from objects, adds CLIP~\cite{radford2021learning} features to voxels on that basis, and maps the feature field to physical properties through a U-Net~\cite{unet}. NeRF2Physics and PUGS render the object from multiple views, then feed the images into a VLM to infer possible physical properties.

\noindent\textbf{Evaluation Metric.}
We apply a logarithmic transformation to $E$ and $\rho$ on the obtained results, i.e., we evaluate the numerical errors of $[\log{E},\,\nu,\,\log\rho]$, as well as the material model prediction accuracy. For fairness, all methods are evaluated in the same six materials, since material types are used for simulation. For LLM-based baselines, NeRF2Physics~\cite{zhai2024physical} and ~\cite{shuai2025pugszeroshotphysicalunderstanding}, we add the simulation requirement and six categories to the prompt and restrict their outputs. Since Pixie~\cite{le2025pixie} already predicts five overlapping simulation materials, to avoid disadvantaging Pixie, we adopt a conservative upper-bound mapping: if GT is plastic, we additionally accept its closest available materials, elastic/metal/rigid, because Pixie does not contain plastic. For evaluation, we randomly select 100 samples from the test set that are not involved in the training and map the materials predicted by other methods to our material models.

\begin{table}[t]
\centering
\small
\setlength{\tabcolsep}{3.8pt}
\caption{Quantitative comparison of material model prediction and physical property errors across baselines.}
\label{tab:phys_results}
\begin{tabular}{lcccc}
\hline
Method & Mat.Acc.$\uparrow$ & log$E$~err.$\downarrow$ & $\nu$~err.$\downarrow$ & log$\rho$~err.$\downarrow$ \\
\hline
NeRF2Physics & 0.628 & 2.033 & 0.064 & 0.521 \\
PUGS          & 0.283 & 2.778 & 0.076 & 0.627 \\
Pixie         & 0.349 & 4.129 & 0.103 & 0.848 \\
\textbf{Ours} & \textbf{0.789} & \textbf{0.661} & \textbf{0.061} & \textbf{0.249} \\
\hline
\end{tabular}
\end{table}
\begin{figure}[t]
    \centering
    \includegraphics[width=1\textwidth]{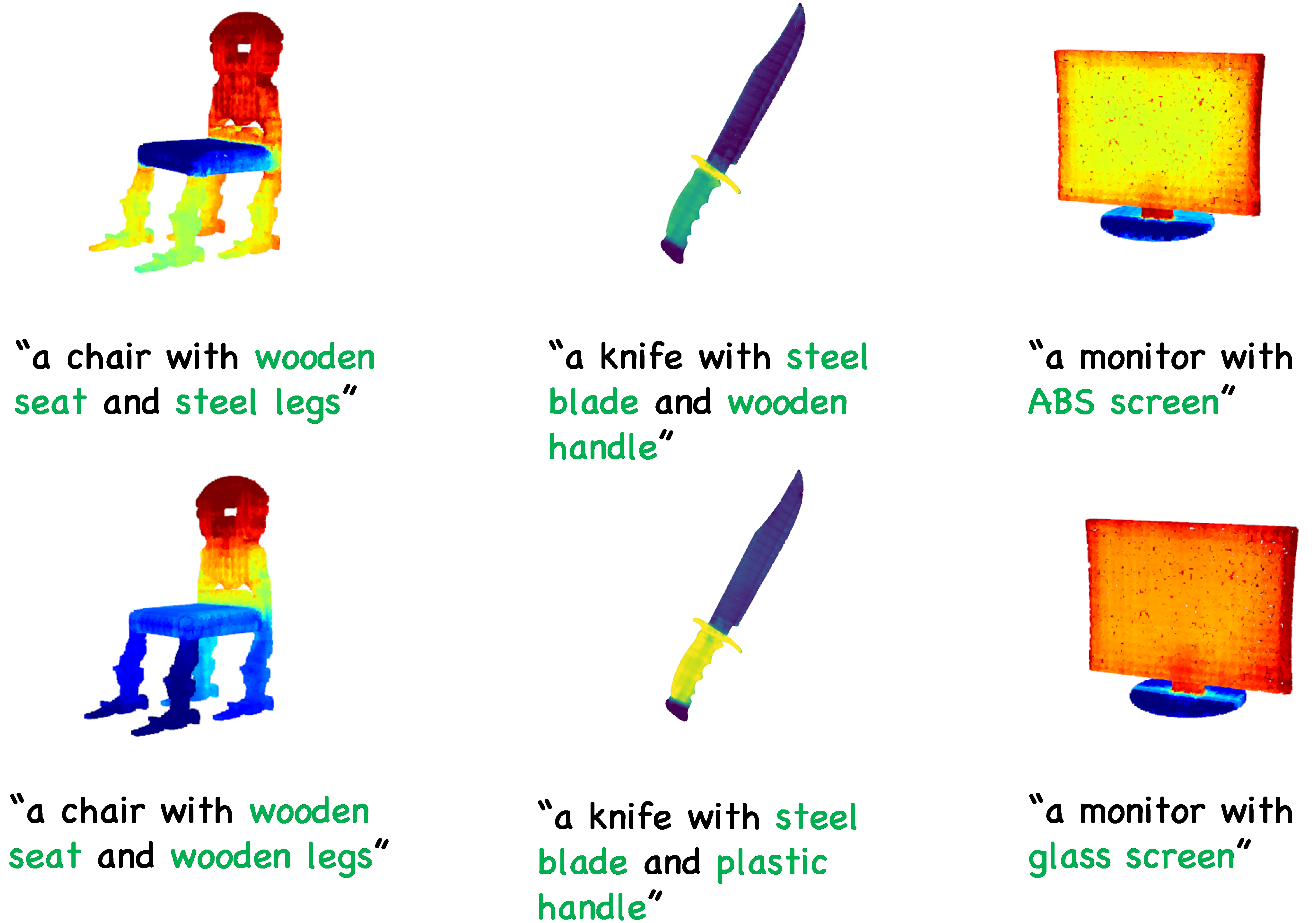}
\caption{Our model can achieve part-level physical property controllable prediction through text condition.}
    \label{fig:phys}
\end{figure}
\begin{figure*}[t]
    \centering
    \includegraphics[width=1\textwidth]{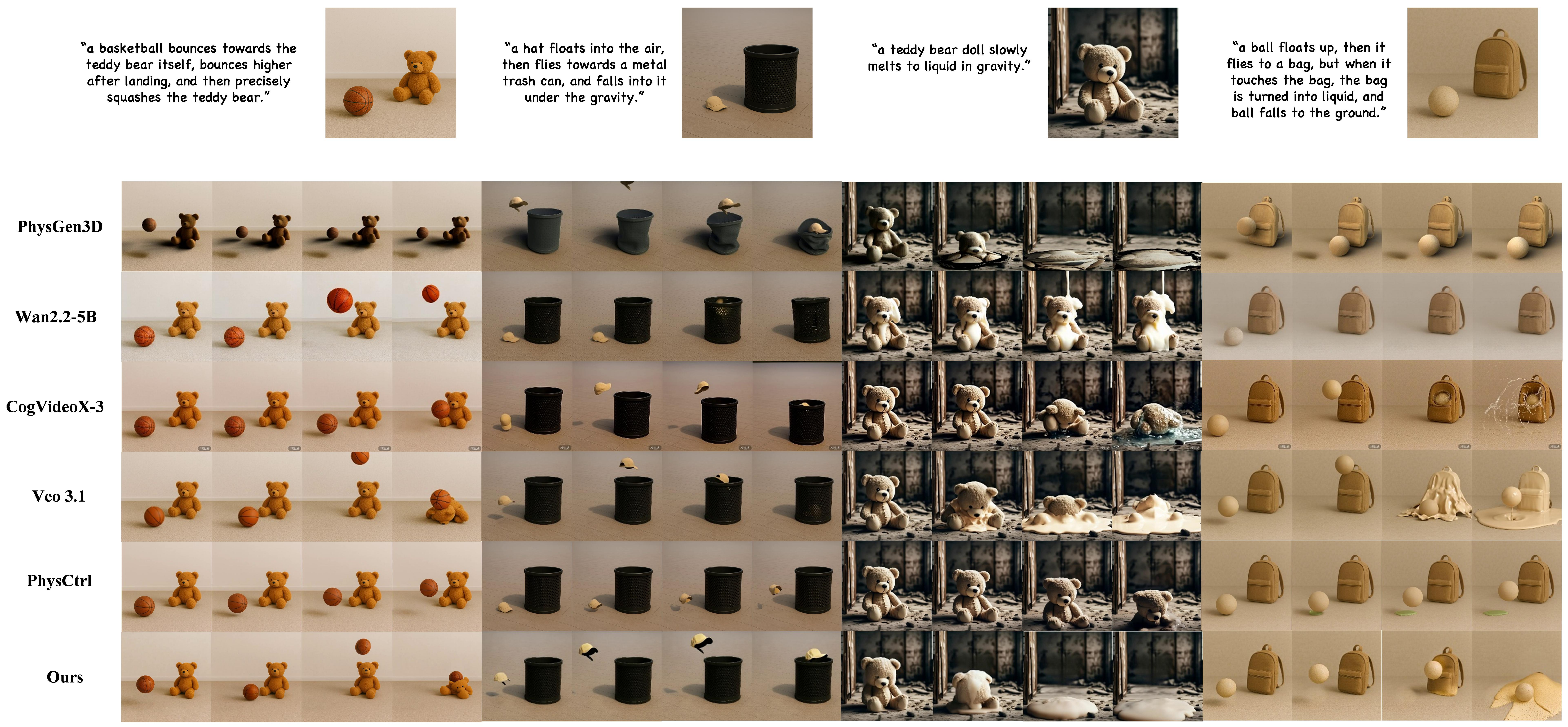}
\caption{Qualitative comparison between PhysChoreo and existing image-to-video generation models.}
    \label{fig:compare}
\end{figure*}
\begin{table}[t]
    \centering
    \begin{minipage}[t]{0.49\linewidth}
        \centering
        \caption{User study results comparing videos from all methods(\%).}
        \label{tab:user_study}
        \resizebox{\linewidth}{!}{%
        \begin{tabular}{lcccc}
            \hline
            Method & SA~($\uparrow$) & PC~($\uparrow$) & VQ~($\uparrow$) & Total~($\uparrow$) \\
            \hline
            PhysGen3D     & 7.7  & 13.0 & 6.7  & 10.42 \\
            PhysCtrl    & 8.5  & 17.0 & 15.2  & 14.1 \\
            Wan2.2-5B     & 3.0  & 4.0  & 6.6  & 4.6  \\
            CogVideoX-3   & 3.5  & 6.4  & 8.2  & 6.0  \\
            Veo~3.1       & 9.9 & 17.7 & 20.5 & 16.1 \\
            \textbf{Ours} & \textbf{67.4} & \textbf{41.9} & \textbf{42.8} & \textbf{50.1} \\
            \hline
        \end{tabular}%
        }
    \end{minipage}\hfill
    \begin{minipage}[t]{0.49\linewidth}
        \centering
        \caption{Quantitative video comparison under VLM evaluation.}
        \label{tab:vlm_eval}
        \resizebox{\linewidth}{!}{%
        \begin{tabular}{lcccc}
            \hline
            Method & SA~($\uparrow$) & PC~($\uparrow$) & VQ~($\uparrow$) & AVG~($\uparrow$) \\
            \hline
            PhysGen3D     & 2.30 & 2.10 & 3.50 & 2.63 \\
            PhysCtrl    & 3.75  & 4.25 & 4.30  & 4.1 \\
            Wan2.2-5B     & 1.75 & 1.70 & 4.20 & 2.55 \\
            CogVideoX-3   & 2.40 & 2.55 & 4.15 & 3.04 \\
            Veo~3.1       & 4.10 & 4.20 & \textbf{4.90} & 4.40 \\
            \textbf{Ours} & \textbf{4.70} & \textbf{4.55} & 4.75 & \textbf{4.67}  \\
            \hline
        \end{tabular}%
        }
    \end{minipage}
\end{table}
\begin{figure}[t]
    \centering
    \includegraphics[width=1\textwidth]{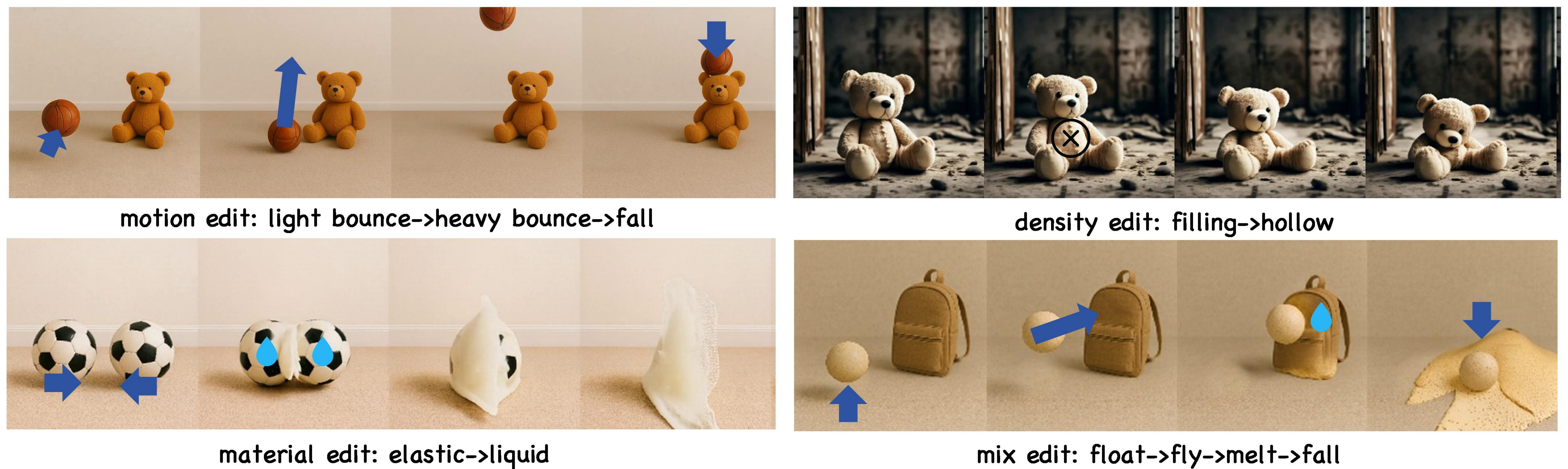}
\caption{PhysChoreo can generate physically realistic and visually appealing videos by changing various physical properties during the runtime.}
    \label{fig:video}
\end{figure}
\begin{table}[t]
    \centering
    \small
    \caption{Ablation study on different components of our framework. }
    \label{tab:ablation}
    \resizebox{1\linewidth}{!}{%
    \begin{tabular}{c|cccc|ccc}
        \hline
        Method & Cross Attn. & Dual Stage. & Assignment & Seg.~Prior & Mat.~Acc.~($\uparrow$) & Total~Err.~($\downarrow$) &Iter.~($\downarrow$) \\
        \hline
        (a) &  &  &  &  & 0.5507 & 3.4230 & \textbf{10} \\
        (b) & \checkmark &  &  &  & 0.7393 & 0.9789 & 35 \\
        (c) & \checkmark & \checkmark &  &  & 0.8077 & 0.5801 & 36 \\
        (d) & \checkmark & \checkmark & \checkmark &  & 0.8372 & 0.5046 & 26 \\
        Ours & \checkmark & \checkmark & \checkmark & \checkmark &\textbf{0.8605} & \textbf{0.3318} & 41 \\
        \hline
    \end{tabular}
    }%
\end{table}
\begin{table}[t]
    \centering
    \small
    \caption{Ablation study on our loss components.}
    \label{tab:loss_ablation}
    \begin{tabular}{lccc}
        \hline
        Method & Mat.~Acc.~($\uparrow$) & Total~Err.~($\downarrow$) & Iter.~($\downarrow$) \\
        \hline
        w/o $\mathcal{L}_{\text{assign}}$   & 0.8534 & 0.3753 & 50 \\
        w/o $\mathcal{L}_{\text{smooth}}$   & 0.8578 & 0.3579 & \textbf{37} \\
        w/o $\mathcal{L}_{\text{con}}$ & 0.8310 & 0.3451 & 38 \\
        \textbf{Full method} & \textbf{0.8605} & \textbf{0.3318} & 41 \\
        \hline
    \end{tabular}
\end{table}
\begin{figure}[t]
    \centering
    \includegraphics[width=1\linewidth]{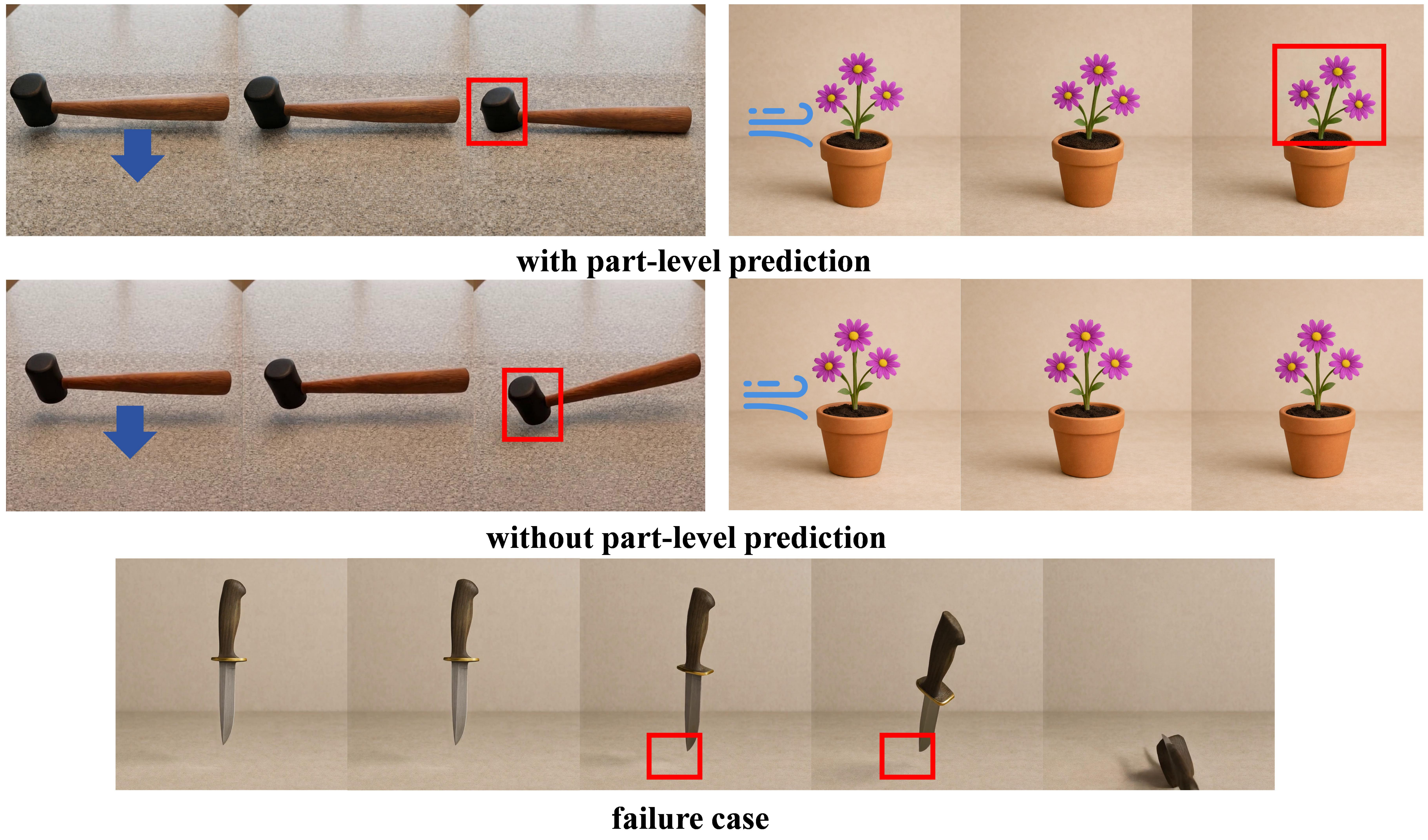}
    \caption{Qualitative results of part-level ablation and failure case.}
    \label{fig:abvideo}
\end{figure}

\noindent\textbf{Results.}
Table~\ref{tab:phys_results} shows the comparison results between our method and baselines. Our method achieves the best performance on all metrics. Since other methods predict fixed material fields while we can learn text- and part-aware distribution, as shown in Fig.~\ref{fig:phys}, our method can use text to control the physical properties of specific parts, providing preliminary controllability for subsequent simulation.

\subsection{Image-to-Video Generation}
\noindent\textbf{Baselines.}
PhysChoreo uses the simulation results to guide video generation. To thoroughly assess the quality of generation, we compare our method with two physics-based methods, PhysGen3D~\cite{chen2025physgen3d} and PhysCtrl~\cite{physctrl2025}, two open-source generative video models, Wan2.2-5B~\cite{wan2025wan} and CogVideoX-3~\cite{yang2024cogvideox}, and the state-of-the-art closed-source video model, Veo 3.1~\cite{deepmind2025veo}.

\noindent\textbf{Evaluation Metric.}
Inspired by VideoPhy~\cite{bansal2024videophy}, we evaluate 10 generated videos using three metrics: (1) Physical Commonsense (PC): whether the motion of the objects follows physically plausible deformations and dynamics; (2) Semantic Alignment (SA): the degree of match between the video content and motion and the text content, with a particular focus on the alignment of temporal instructions; (3) Visual Quality (VQ): the detailed visual quality of the video. Each metric is scored on a 5-point scale. Then, based on these metrics, Gemini-2.5-Pro~\cite{team2023gemini} scores each video and 31 real users choose the preferred selection of each case.

\noindent\textbf{Results.}
We report the VLM scores in Table~\ref{tab:vlm_eval} and 642 valid selection results in Table~\ref{tab:user_study}. Thanks to physics-based simulation, our generated videos achieve the best PC.
Since we explicitly predict the dynamics of objects, PhysChoreo achieves a significant lead on SA. While the visual quality of our generated videos is not as good as Veo3.1's, our method achieves the best average score. Qualitative results in Fig.~\ref{fig:compare} are basically consistent with our evaluation scores. Furthermore, we demonstrate in Fig.~\ref{fig:video} that, by manipulating the physical properties in the scene, PhysChoreo can generate various behaviors like liquefying upon collision, collapsing upon landing, and counter-intuitive bounces.

\subsection{Ablation Studies}
\label{sec:ablation}
\noindent\textbf{Model Design.} To evaluate the effectiveness of our modules, we conduct ablation experiments in Table~\ref{tab:ablation}(iteration denotes the early stopping epoch). Our baseline is to directly predict physical properties through a transformer model. Based on this, we gradually incorporate cross-attention, soft assignment, and segmentation priors. In (a), the error is too large to accurately predict the physical properties. In (b) and (c), text features significantly improve performance. In (d), soft assignment that explicitly enhances the injection of text features, improves efficiency. Finally, the segmentation prior increases the training time for the added dimensions but gets the best performance.

\noindent\textbf{Physics-based Supervision.} We conduct ablation experiments on loss functions. In Table~\ref{tab:loss_ablation}, without text-part assignment supervision $\mathcal{L}_{\text{assign}}$, performance decreases slightly, but iterations increase markedly. Without $\mathcal{L}_{\text{smooth}}$, performance declines slightly for the high-frequency noise within the parts' attributes. Removing contrast supervision $\mathcal{L}_{\text{con}}$ achieves a similar error score, but material accuracy decreases. This is reasonable because smooth supervision improves the prediction of boundary points.

\noindent\textbf{Part-level Knowledge.} To assess the effectiveness of part-level knowledge within PhysChoreo, we conduct several experiments and add comparison videos with manually specified physics. These oracle global settings share the same simulator and video pipeline, but miss part-specific and temporally edited dynamics. As shown in Fig.~\ref{fig:abvideo}, when directly using global properties instead of part-level prediction, the results fail to maintain correct part-level dynamics. Finally, with part-level properties predicted by PhysChoreo, our results exhibit more distinct part-level responses.

\subsection{Failure case}
Common failure cases occur in the rendering stage. Our simulation ensures stable physical trajectories, but current video models occasionally fail to handle correct light and shadow details in dynamic motions, leading to visible artifacts in fine-grained appearance, which is shown in Fig.~\ref{fig:abvideo}.
\section{Conclusion}
We introduce PhysChoreo, demonstrating excellent performance in reconstructing physical properties and generating rich dynamic videos, with the prediction model and solver also having broader applications such as robot simulation.

\noindent\textbf{Limitation.} Our method focuses on independent objects, thus still lacking adequacy for large-scale scenes. Additionally, despite adopting point sampling methods, the internal physical states cannot be precisely predicted. Future work includes addressing these toward broader scenarios.
\section*{Acknowledgements}
This work is supported in part by the National Natural Science Foundation of
China (NSFC) under Grant No. 62371164.
{\small
\bibliographystyle{splncs04}
\bibliography{main}
}

\end{document}